# Heterogeneous Noisy Short Signal Camouflage inMulti-Domain Environment Decision-Making*

Piyush K. Sharma

Army Research Laboratory
Adelphi, MD 20783, USA

**Abstract.** Data transmission between two or more digital devices in industry and government demands secure and agile technology. Digital information distribution often requires deployment of *Internet of Things (IoT)* devices and Data Fusion techniques which have also gained popularity in both, civilian and military environments, such as, emergence of *Smart Cities* and *Internet of Battle eld Things (IoBT)*. This usually requires capturing and consolidating data from multiple sources. Because datasets do not necessarily originate from identical sensors, fused data typically results in a complex *Big Data* problem. Due to potentially sensitive nature of IoT datasets, *Blockchain* technology is used to facilitate secure sharing of IoT datasets, which allows digital information to be distributed, but not copied. However, blockchain has several limitationsrelated to complexity, scalability, and excessive energy consumption. We propose an approach to hide information (sensor signal) by transforming it to an image or an audio signal. In one of the latest attempts to the military modernization, we investigate sensor fusion approach by investigating the challenges of enabling an intelligent identification and detection operation and demonstrates the feasibility of the proposed *Deep Learning* and *Anomaly Detection* models that can support future application for specific *hand gesture* alert system from wearable devices.

## 1 Introduction

Advances in technology have contributed to the growing use of laptops, smart-phones, and tablets in government and industry globally. Nascent 5G network enables mobile computing more effectively [13]. Increasing amount of information ow are impacting daily life of individuals, business conducted across industry, and government operations [3]. Expansion of digital information introduces several challenges related to complexity of data management, storage, privacy, processing and transfer. These challenges are also seen in utilizing information for decision-making. In order to leverage this abundant, raw, digital information, military must keep abreast with advanced data-driven technology to meet mission requirements.

---
* This work was supported by the U.S. Army Research Laboratory (ARL).

Modern military's Multi Domain operations (MDO) can potentially utilize Internet of Things (IoT) technology connecting land, water, air, space and cyberspace in a cohesive network [1,2,10]. IoT devices (sensors, actuators, cameras, etc.) capture and consolidate data from multiple sources allowing the full realization of C2 (Command and Control) system to provide situational awareness [35,38,39]. This combined information from a range of heterogeneous IoT devices can give an edge for potential strategic advantages by increasing situational awareness and risk assessment, reducing response time, and allowing military to take agile, perceptive, resilient, and reliable decisions in the future battle elds.

This study aims to measure the influence different aspects of human state have on how a person makes gestures, and if fusion data approach can improve gesture recognition [22]. Future military combat suits, helmets, weapons, and other equipment will be embedded with sensing and computing devices. These devices not only provide real-time interaction between soldiers and commanding officers in combat zones, but also help understand the psychophysical condition of soldiers. A controlled exploitation of this approach can reduce cognitive load of soldiers, improve decision-making, and establish a better soldier-machine (IoT devices) interaction. Experimental data was collected with the sole purpose of training machine learning algorithms to automatically detect the onset and o set of separate gestures in a series.

Research literature presents examples where multi-sensor data fusion techniques were implemented in different application domains. One paper proposed a smart home system with wearable intelligent technology, artificial intelligence, and multisensor data fusion technology [14]. A 3D gesture recognition algorithm was developed to recognize hand gestures. Another work presented a human activity recognition example using wearable and environmental sensor data fusion approach [28]. Furthermore, examples of Human-Computer-Interaction (HCI) for sign language and gesture recognition have been proposed within a multi- sensor fusion framework [19,20].

In Section 2, we describe challenge and our approach. In Section 3 we define methods and techniques used. In Section 4, we describe data used in experiments. In Section 5, we describe why we choose performed experiments. In Section 6, we explain data encoding approach. In Section 7, we provide a comparative analysis of results obtained. Finally, we conclude our contribution with a summary of results in Section 8.

## 2   Challenge and Proposed Approach

Growing number of cyber-attacks put sensitive information at risk [31]. Information is a strategic asset for government organizations, and they need to protect it for national security [34]. Recently Mirai botnet malware launched a distributed denial of service (DDoS) attack causing much of the internet inaccessible on the U.S. east coast [11]. With the Army's growing dependency on IoT devices and the possible failure of current technology to keep up with the cyber security, C2operations are susceptible to cyber-attack from possible threats capable of compromising IoT devices to exterminate multi-domain operations by injecting false or compromised information [18]. Hackers are developing new variants of IoT-focused malware with alarming regularity. Due to potentially sensitive nature of IoT datasets, Blockchain technology is used to facilitate secure sharing of IoT datasets, which allows digital information to be distributed, but not copied [25]. However, blockchain has several limitations related to complexity, scalability, and excessive energy consumption [26]. The key questions we ask are: How to tailor information for transmission over a compromised network and retain most of it in the process with the goal of enabling decisions? How to model complex systems, detect and understand issues, and improve transmission?

In order to address these questions and communication challenge, we propose a data camouflage approach to modify information (signal) by transforming it to an image or an audio (Section 6) [37,15,46]. Our contributed data encoding approach has multiple advantages; transformed data preserves all information, generated image and audio les are small in size thus can easily be transferred over internet, transformed data is invertible and easier to visualize. Therefore, we can retrieve original signal from encoded data without requiring an encryption key. Moreover, a senseless and unstructured data can be transformed into structured data from which meaningful information can be extracted to be used as actionable insights that enable intelligent military decision-making [35]. Be-cause timing is crucial for mission success; suspicious and potential targets can be identified and neutralized instantly with Edge Computing [12].

## 3   Methods and Techniques Used

We employ classical Machine Learning, Deep Learning and Transfer Learning techniques on data before and after encoding for a comparative analysis [29]. We explore recently introduced information theoretic kernels, Chisini Jensen Shannon Divergence (CJSD) and its metric version, Metric-Chisini Jensen Shannon Divergence (M-CJSD), known for their utility to tease apart data classes with discernible differences in Support Vector Machine (SVM) classification [43], [38], [39], [40], [41], [42]. Finally, we explore a number of novelty detection techniques for their ability to identify anomalous observations which deviate significantly from the majority of data. It is in contrast to classification models which assign predicted classes into groups.

### 3.1   Classification with SVM and CJSD, M-CJSD Kernels

We employ family of CJSD and M-CJSD kernels, and evaluate their performance in SVM for gesture classification. Our kernel function takes input variables from each classification problem's dataset. Because these information theoretic kernels are derived from the mutual information between probability distributions, it is expected that the classifiers will choose the best answer according to the probabilistic model thus improving the classification accuracy.

We estimated probability densities of CJSDs and M-CJSDs using multivariate kernel density estimation (KDE), a nonparametric approach, on each dataset [33,30,45]. This distribution is defined by a smoothing function and a bandwidth value that controls the smoothness of the resulting density curve. Results were validated through 10-fold nested cross-validation on the randomized datasets constructed from binary classification problem. It is important to note that datasets used in experiments for each type of CJSD and M-CJSD kernel (A.M., G.M., H.M.) were randomized using unique random number seeds. For fair comparison, we tested RBF kernels with the same randomized datasets. This resulted in a total of 21 kernel versions (3 for RBF and 18 as explained below).

Let, $P = \{p_i\}_{i=1}^N$, and $Q = \{q_i\}_{i=1}^N$ be two probability distributions, where $p_i$ and $q_i$ are the respective probabilities associated to the $i^{th}$ state (possible values). The family of CJSDs is defined in [43] with the following expression:

$$CJSD(P||Q) = \frac{1}{2}\left[\sum_{i=1}^N p_i log \frac{p_i}{M_i} + \sum_{i=1}^N q_i log \frac{q_i}{M_i}\right] \tag{1}$$

where, $M$ is the mid-point of distributions $p_i$ and $q_i$. If $M$ is the arithmetic mean, equation (1) becomes Jensen-Shannon divergence (JSD). This modified CJSD is useful when distributions are similar and it is hard to distinguish them. A further modification was proposed in [39] which extends CJSDs to exploit the metric properties of JSD. Metric-Chisini-Jensen-Shannon divergences (M-CJSD) is defined as:

$$M - CJSD(P||Q) = \sqrt{CJSD(P||Q)} \tag{2}$$

Some M-CJSD based kernels are:

$$K_{Amplified} = M - CJSD(P||Q)e^{-\frac{|x_i - x_j|^2}{2\sigma^2}} \tag{3}$$

$$K_{Scaled} = e^{-M-CJSD(P||Q)\frac{|x_i - x_j|^2}{2\sigma^2}} \tag{4}$$

$$K_{Amplified-Scaled} = M - CJSD(P||Q)e^{-M-CJSD(P||Q)\frac{|x_i - x_j|^2}{2\sigma^2}} \tag{5}$$

Likewise, replacing M-CJSD with CJSD in (3), (4) and (5), we get CJSD based kernels. In this paper, we study these kernels for *Arithmetic, Geometric,* and *Harmonic* (AM, GM, HM) means. Therefore, we employ AM, GM, HM for *Amplified, Scaled,* and *Amplified-Scaled* kernel versions for both M-CJSD and CJSD. This results in a total of 18 kernels.

## 3.2   Deep Learning

For the initial set of experiments, we employ *transfer learning* approach. This requires selection of a suitable a pre-trained model. For this purpose, we tried different pre-trained models from the list of available models for image classification with weights trained on *ImageNet* (Xception, VGG16, VGG19, ResNet, ResNetV2, InceptionV3, etc.) [9]. Our experiments show that all models give similar classification results, therefore, we report results only for ResNet50. A summary of the trainable parameters of the model is provided in Table 1. We ne-tune the model by tweaking its parameters for our data and freeze all the layers except the last 4 convolutional layers which we use for training. Fine-tuning avoids limitations of model by not training from scratch on small data and saving training time (because less parameters will be updated in training). We evaluate model performance by computing validation accuracy and validation loss.

For DNN hyperparameter tuning, we searched over a large grid with *Batch Size, Epochs, Learn Rate, Optimizer, Momentum, Initial Mode, Activation, Dropout Rate, Weight Constraint,* and *Neurons.* DNN and radial basis function

Table 1: Model summary of trainable parameters

| Layer (type) | Output Shape | Param # |
|---|---|---|
| resnet50 (Model) | (*None*, 8, 8, 2048) | |
| atten_1 (Flatten) | (*None*, 131072) | 0 |
| dense_1 (Dense) | (*None*, 1024) | 134218752 |
| dropout_1 (Dropout) | (*None*, 1024) | 0 |
| dense_2 (Dense) | (*None*, 2) | 2050 |

Total params:          157, 808, 514
Trainable params:     135, 275, 522
Non-trainable params: 22, 532, 992

(RBF) kernel are used as the performance benchmark for comparison with CJSD and M-CJSD kernels. We report the classification results with sample mean and standard error in error bar plots (Subsection 7.1).

### 3.3  Novelty Detection

Our goal is to increase confidence in decision-making in military operations using our model's ability to detect gestures correctly for soldiers being able to decide whether a new observation belongs to the same distribution as existing observations (inlier), or should be considered as different (outlier). Therefore, for the final set of experiments, we prepare our training data which is not polluted by outliers and detect whether a new observation is an outlier. In this context an outlier is also called a novelty. We perform anomaly detection by training the model only on the positive (known) class dataset and predicting negative (unseen) classes. Out of many available algorithms for anomaly detection, we choose One-class SVMs [5], Isolation Forests [23] and Gaussian Mixture Model (GMM) [32]. We use isotonic regression to convert the output of the GMM to a probability score. We train our models on No-Gesture as a positive class and use it to predict Gesture as a negative class.

### 3.4  Estimation of Model Performance

We evaluate models' performances by comparing their respective accuracy, precision, sensitivity (true positive rate) and specificity (true negative rate) along with confusion matrices, ROC and Precision-Recall curves. We used LIBSVM library in MATLAB to employ aforementioned kernels [4]. Additionally, we provide a run time comparison for all models and summarize results in Table 5.

For all other experiments and data preprocessing, we used the latest versions of Python and Keras framework with TensorFlow as a backend [6]. Our computing system consisted of 128 GB RAM for CPU, and NVIDIA Quadro P3200 6144MB - Memory Type: GDDR5 (Samsung).

## 4    The Data

Military operation often require communication with minimum sound and therefore use tactical hand gestures for communication and coordination. As we set out to investigate how soldiers' can increase situational awareness by using gestures in military operations, our data came from physiological experimentation, however, this paper is focused on use by soldiers versus medical doctors [7].

Gesture datasets were collected by a *Myo* armband, made by *Thalamic Labs* (Figure 2). *Myo* armband is embedded with multi-sensory devices, therefore, it can potentially operate in IoT and military operations. Aggregated data is noisy has features at different scale (Figure 1). Feature values corresponding to gestures are plotted on vertical axis. Square wave wrapping around the signal shows beginning and the end of a gesture being made. So, the region outside the square wave represents features values when no gesture was made. These feature values seem to be on top of each other indicating data nuance challenge. Section 5 explains data collection process and our how we use it in gesture recognition task.

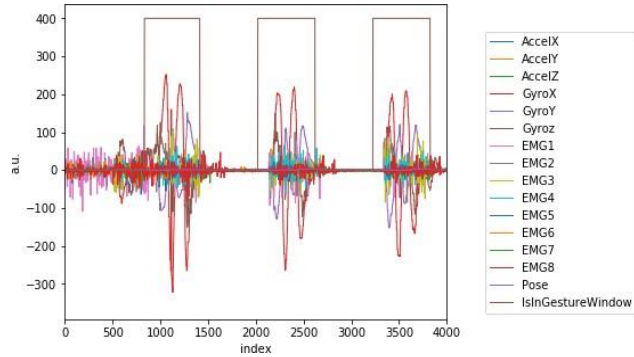

Fig. 1: Illustration of data originated from the Myo device and Unity gesture experiment program. Square wave wraps cleanly around the rest of the data, shown as noisier channels within the square wave.

## 5    Experiments

The goal was to develop an algorithm that can automatically provide the likely start and stop indices of each gesture segment enabling the parsing (and labeling) of data from the actual experiment for further analysis and modeling. Subjects wore the *Myo* armband on their left arm and performed a series of NATO gestures one after the next according to a display on a screen in front of them. The display was run in the *Unity* game engine. Subjects relaxed their arm in between gestures. Timing information of the visual display and data from the Myo armband were recorded and synchronized through custom code using Lab-StreamingLayer (LSL) [17]. This synchronization produced a data frame (Table 2), that enabled analyses of all data within the same time-space (Figure 1). Detailed description of the experimental design is given in [21].

## 5.1 Experimental Setup

The device contains an *inertial measurement unit (IMU)* and *electromyography (EMG)* sensors to measure changes along the up-down, left-right, and front-back axes and changes in muscle activity, respectively. These measurements are based on electrical stimulation created by flexing fore-arm muscles coupled with a 9-axis IMU, which in fact is comprised of 3-axis Gyroscope (Gyro), a 3-axis Accelerometer (Acc), and a 3-axis Magnetometer to (MAG) track arm movement.

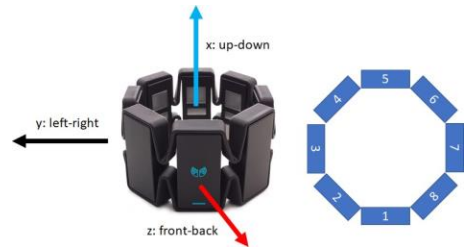

Fig. 2: The Myo armband used in actual experimentation with associated IMU axes and EMG sensor positions.

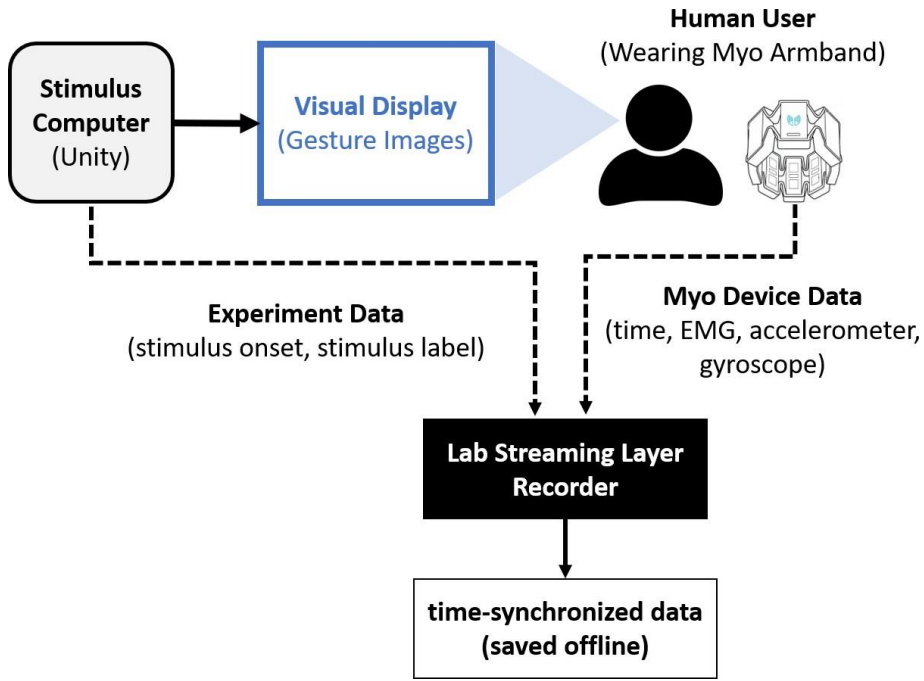

Fig. 3: Experimental Design.

## 5.2   Gesture Myo Data

We created three different datasets by pairing Myo armband sensors. For the first pair, we consolidated 3-axis accelerometer, 3-axis gyroscope from IMU, and 8-axis EMG giving a fused signal of length 14. For the second pair, we consolidated only 3-axis accelerometer and 3-axis gyroscope from IMU giving a fused signal of length 6. For the third pair, we considered only 8-axis EMG. Our data consisted of a total 38507 instances, with 24845 instances of no gesture, and 13662 instances of gestures. We preprocess data to drop instances and features with NaN and infinite entries. We also drop duplicated instances and provides a summary of each dataset after preprocessing step. In our experiment each instance represented a signal and we used the channels at each axis as a feature. Therefore, our data instances were vector valued in 14, 6, and 8 dimensions respectively.

Table 2: Data summary

| Sensor Name | Channel |
|---|---|
| Electromyogram (EMG) | $1 - 8$ |
| Accelerometer | $9 - 11$ |
| Gyroscope | $12 - 14$ |
| Myo Pose Classification | 15 |
| Class Label (Gesture or No-Gesture) | 16 |

Table 3: Number of *Gesture* and *No-Gesture* instances after data preprocessing.

| Data Type | Sensors Included | Instances × Features | Gesture Instances | No-Gesture Instance |
|---|---|---|---|---|
| Signal (Before Encoding) | Acc+Gyro+EMG | 38507 × 14 | 13662 | 24845 |
| | Acc+Gyro | 38507 × 6 | 13662 | 24845 |
| | EMG | 38507 × 8 | 13662 | 24845 |
| Image Features (After Encoding) | Acc+Gyro+EMG | 36139 × 512 | 13097 | 23042 |
| | Acc+Gyro | 30576 × 288 | 9798 | 20778 |
| | EMG | 19549 × 288 | 9634 | 9915 |
| Audio Features (After Encoding) | Acc+Gyro+EMG | 38507 × 20 | 13662 | 24845 |
| | Acc+Gyro | 3302 × 20 | 3269 | 33 |
| | EMG | 5504 × 20 | 5162 | 342 |

We explore machine and deep learning models which we benchmark against anomaly (novelty) detection models (Subsection 3.3). We provide empirical validation of applied techniques on datasets before and after encoding. Therefore,we validate our models on fused sensor data (namely, *signal data*), on images obtained with *signal-image* encoding, on image and audio features (Section 6). For all experiments, we build a binary classification problem with 75:25 percent training and validation data splits. Data instances were labeled as *Gesture* and *No-Gesture*. For transfer learning, we did 80:20 percent data split.

Because one research objective is to find the suitable pairing for sensor data fusion which gives the best output by preserving patterns of the encoded in- formation, we analyze aforementioned data types (signal, image, audio) on Accelerometer (Acc), Gyroscope (Gyro), EMG pairs. Therefore, our experiments consist of 9 datasets, plus one additional 10th (*Acc+Gyro+EMG*) dataset of encoded images used in transfer learning model.

Principal Component Analysis (PCA) is often used to visualize a high dimensional data by projecting it onto a low dimensional subspace [16,36,44].

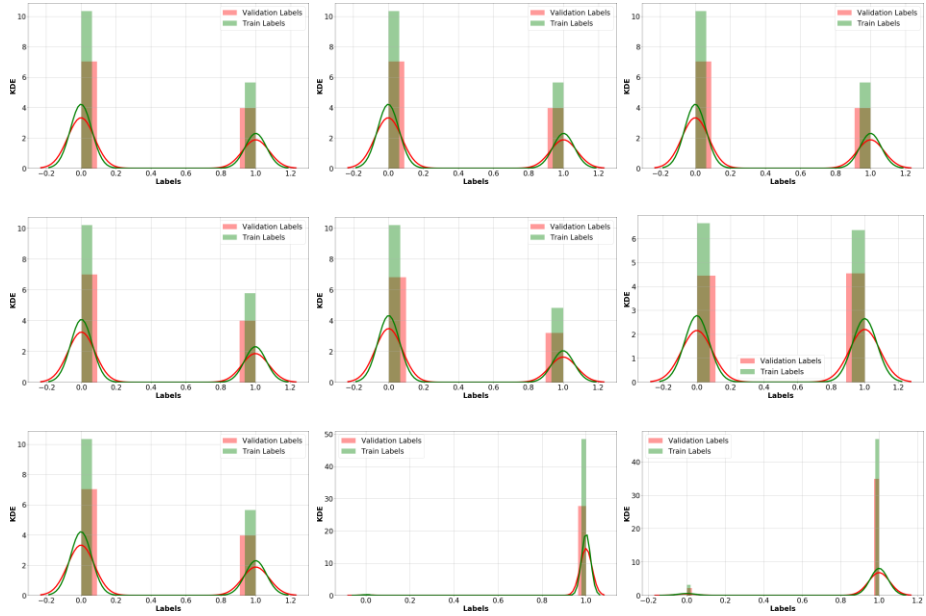

Fig. 4: Distribution of *Train* and *Validation* labels shows high class imbalance for datasets (Left to Right)*Acc+Gyro+EMG*, *Acc+Gyro* and *EMG* for data type (Top to Bottom) signal, image and audio.

A visualization of audio data is shown in Figure 6. Moreover, summary for each dataset after preprocessing step is provided in Table 3. For image features, we selected the top 30 principal components that explained over 99% variance.

## 6    Data Encoding

Because we are dealing with short signals, in order to produce images, we make them of lengths 16, 9 and 9 by adding 2, 3, and 1 zeros at the end of the signals respectively. This is called *Zero Padding* which is useful in many applications allowing us to increase the frequency resolution arbitrarily. These padded signals can be converted to images of size 4 x 4, 3 x 3, and 3 x 3 respectively. Next, we scale each signal between 0-255, and then use *Python Imaging Library (PIL, aka Pillow)* library to save encoded signal as an image.

For computing texture features, we use GIST descriptor which uses a wavelet decomposition of an image [27]. An image has 1 GIST descriptor of length 512. Thus, each instance represents a single GIST descriptor of 512 dimensions. Data thus obtained poses challenges with high dimensionality (Figure 5).

Similarly, for signal to audio encoding, we pad signal with zeros and use Python's standard library with *wave* module providing an easy interface to the audio WAV format. This is like treating a signal as if the short burst is followed by silence and does not impact our data significantly. However, for a long zero padding a reasonable approximation of the actual note is required. We obtained audio les of length 9, 4 and 4 seconds for *datasets Acc + Gyro + EMG, Acc + Gyro* and *EMG* respectively.

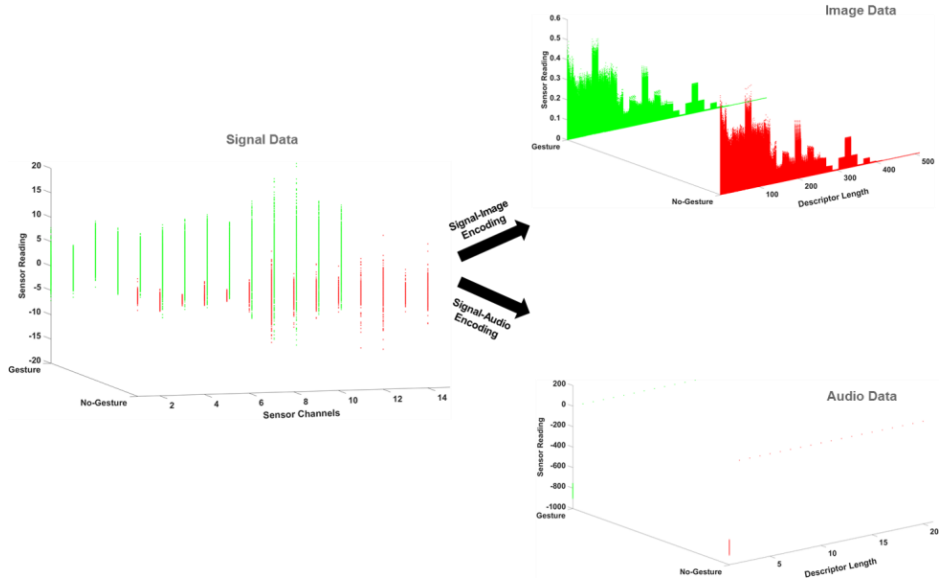

Fig. 5: Sensor to image and audio data encoding. Green and red colors correspond to *Gesture* and *No-Gesture* instances. For audio, colors appear very faint on axes.

   For computing audio features, we use MFCC descriptor [8], [24]. Audio signals can geometrically be resented in 3D space with time, amplitude and frequency. We extract 1 MFCC descriptor of length 20 from each audio clip. Thus, each instance represents a single MFCC descriptor of 20 dimensions.

## 7   Results and Discussion

Here we report empirical validation results for signal, image and audio datasets with aforementioned classical and deep learning approaches. We provide a detailed comparative analysis of models' classification performances before and after camouflage with *signal-image* and *signal-audio* encoding schemes.

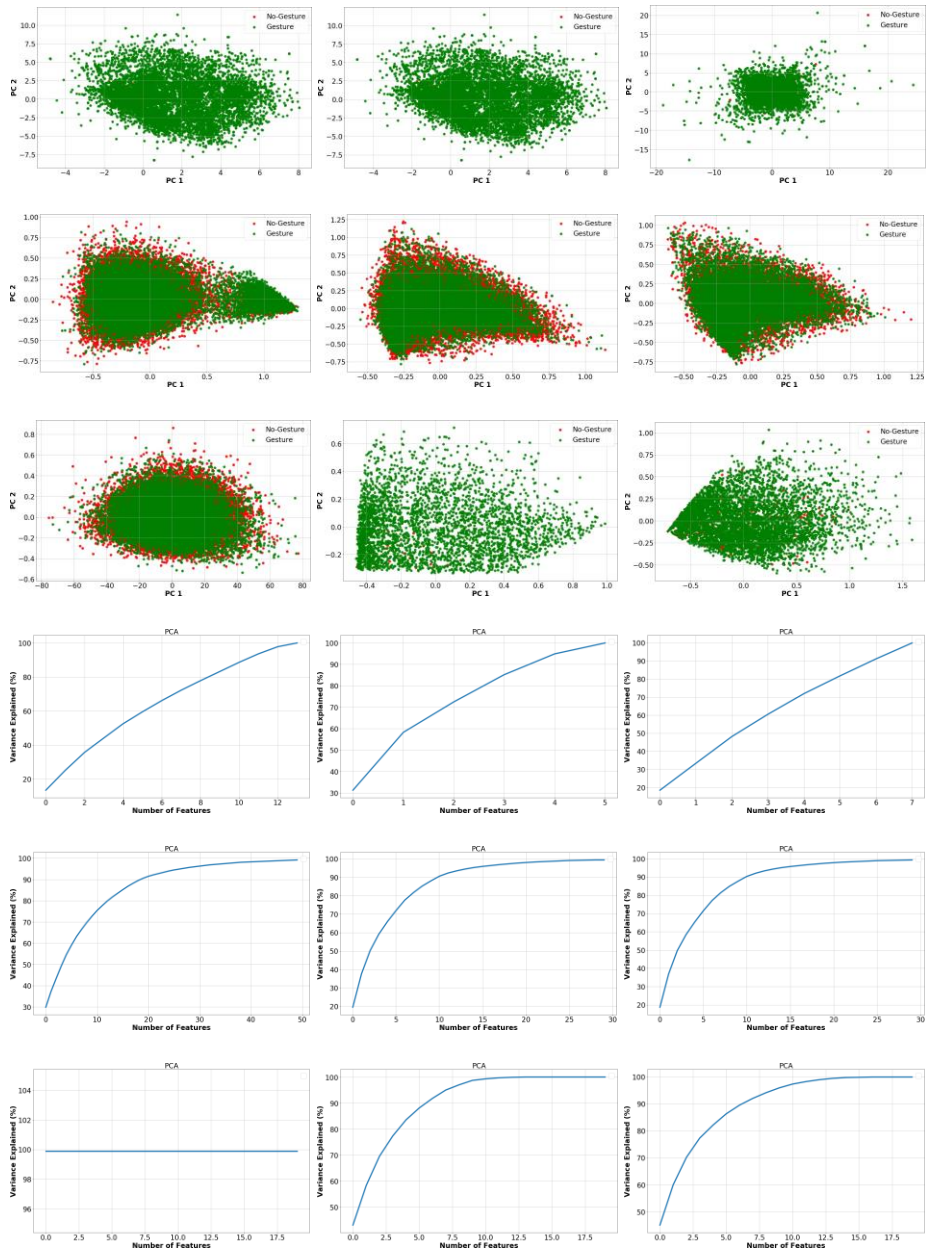

Fig. 6: (First Three Rows) PCA scatter plots show high class similarity between Gesture (green) and No-Gesture (red) classes. (Last Three Rows) Corresponding *Principal Components vs. Explained Variances* plots for datasets (Left to Right)*Acc+Gyro+EMG, Acc+Gyro* and *EMG* for data type (Top to Bottom) signal, image and audio.

## 7.1 Classification (Before and After Encoding)

All CJSD and M-CJSD kernels in SVM classification achieved 99% accuracy on audio data for *Acc+Gyro* and 93.79% for EMG outperforming all signal and image data pairs with statistical significance at the 95% confidence level. Moreover, all kernels except *Sca* and *Amp_Sca* versions gave higher classification accuracy on signal data than on image data for *Acc+Gyro* with statistical significance at the 95% confidence level. Also, all kernels except *Sca* versions gave higher classification accuracy on signal data than on audio data for *Acc+Gyro+EMG* with statistical significance at the 95% confidence level (Figure 7). We use the overlap in confidence intervals to check the statistical significance. As the intervals do not overlap, there is at least 95% confidence (with *p*-value at the $p < 0.05$ level of significance). Our analysis shows that the best classification results are obtained on audio data for *Acc+Gyro*.

On the other hand, run-time comparison shows that DNN outperforms CJSD and M-CJSD based kernels significantly on signal and image datasets. However, on audio data for *Acc+Gyro* and *EMG*, comparison shows similar run time performance. Because we tested 21 kernels, run time can be computed by dividing total time with 21 from Table 5.

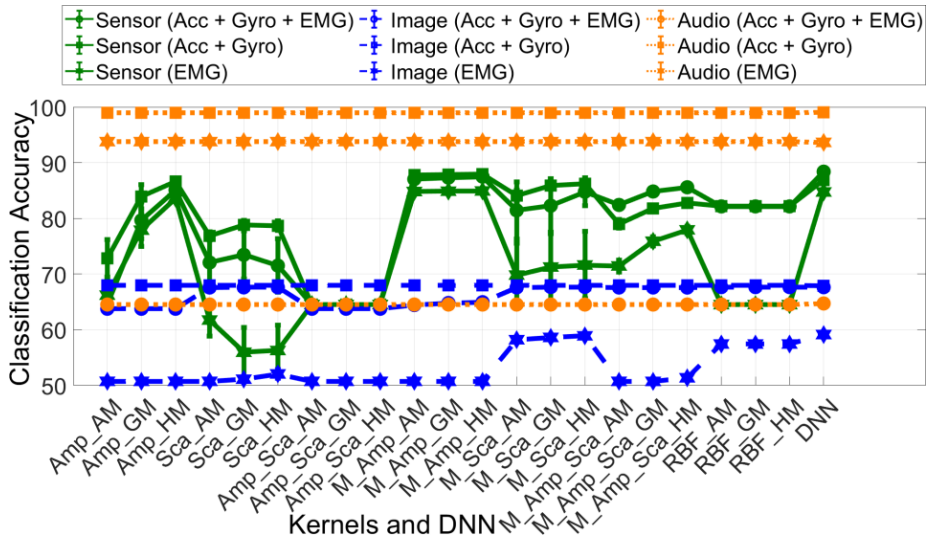

Fig. 7: Error bars show accuracy achieved by CJSD, M-CJSD, RBF kernels in SVM classification and DNN for sensor (green), GIST with image resized to 256 x 256 (blue), and GIST with original 4 x 4 image (orange). Results are reported for 3 randomized data versions (*AM, GM, HM*) for CJSDs and M-CJSDs. For signal data (green), notice that DNN and all metric CJSD kernel version except its scaled version outperform other kernels achieving around 88% average accuracy. RBF performs the same for each randomized data version achieving around 82% average accuracy. For GIST features (orange and blue), the highest accuracy achieved by some kernels is around 67% and for DNN it is around 66%. Finally, notice the high standard error for DNN and some kernels.

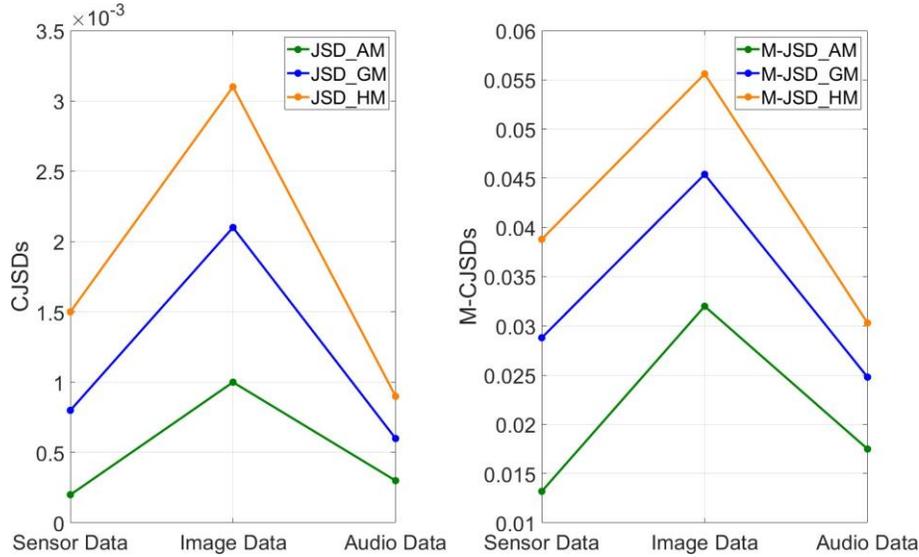

Fig. 8: CJSDs & M-CJSDs separate the probability distributions of *Gesture & No-Gesture* instances for each dataset in the following order: $JSD_{HM} > JSD_{GM} > JSD_{AM}$ and $\sqrt{JSD_{HM}} > \sqrt{JSD_{GM}} > \sqrt{JSD_{AM}}$ satisfying the respective propositions [43,39]. Lower value indicates more while higher value indicates less similarity between the distributions.

Transfer Learning achieved 64.52% average accuracy with 4.84 loss on validation set (Figure 9). DNN achieved maximum classification accuracy of 88.39% on signal data for *Acc+Gyro+EMG*, of 68% on image data for both *Acc+Gyro* and EMG, and of 99% on audio data for *Acc+Gyro*.

A visualization of DNN accuracies for each data type from their respective pots show that our model gives similar results on training and validation datasets. Even though its performance on image data is lower than on signal data, it ts almost similarly on both. However, for *Acc+Gyro+EMG* audio data, model performs poorly on validation dataset, while *EMG* audio data it over ts. This perhaps is due to only a few *No-Gesture* samples in validation dataset (Table 3). It gives the best output on *Acc+Gyro* audio data. Because validation dataset contains only a few No-Gesture samples, we plot ROC-AUC and Precision-Recall curves which con rm DNN performing the best on *Acc+Gyro* audio data. This suggests that for short signals, signal-audio encoding preservesmost data patters and *EMG* is redundant for analysis.

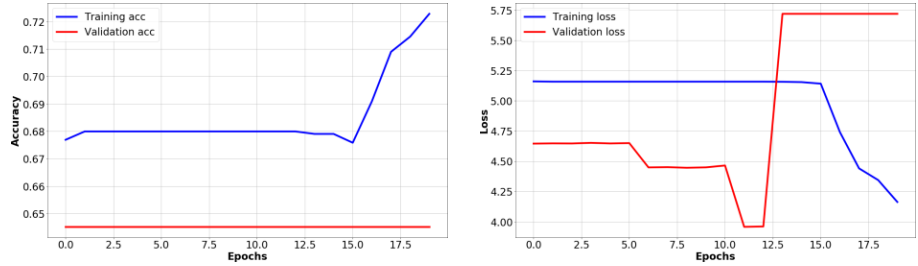

Fig. 9: Transfer Learning Training and Validation Accuracy and Loss.

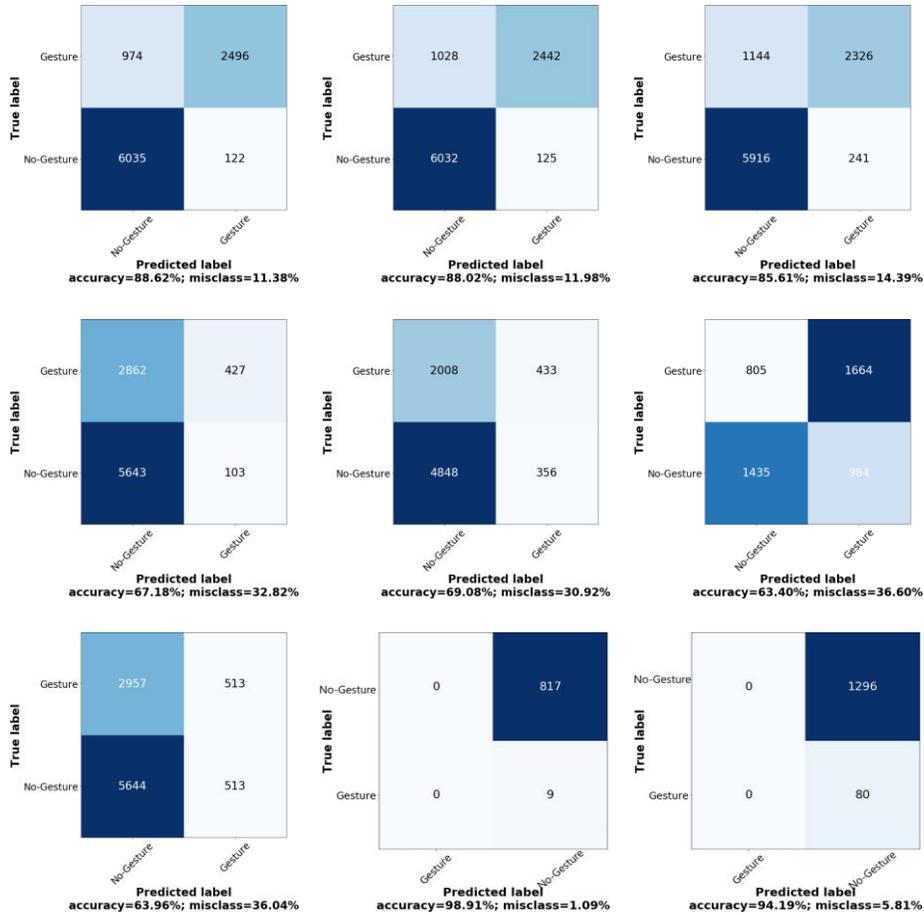

Fig. 10: DNN Confusion matrices trained on 75% *Gesture* and tested on 25% *No-Gesture* instances for datasets (Left to Right) *Acc+Gyro+EMG*, *Acc+Gyro* and *EMG* for data type (Top to Bottom) signal, image and audio respectively.

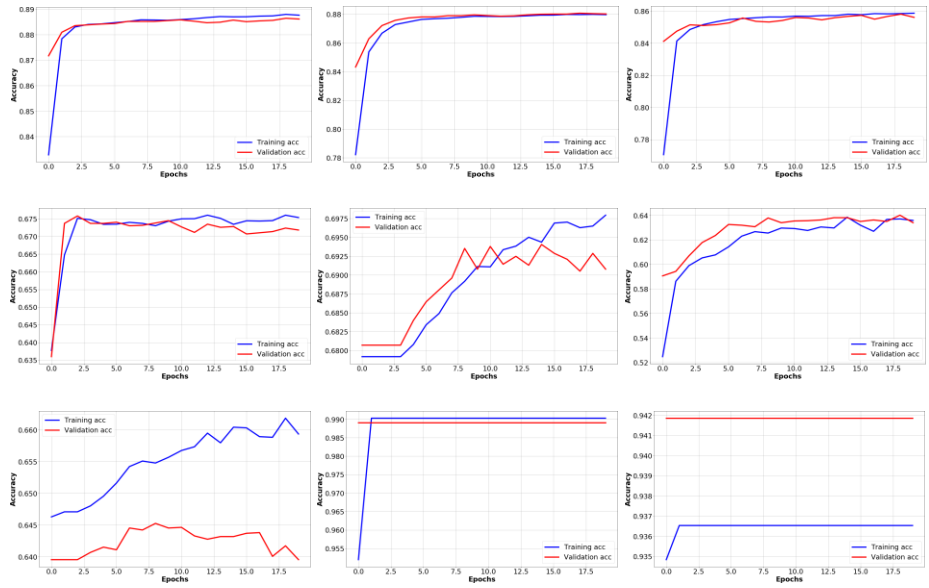

Fig. 11: DNN classification accuracy for datasets (Left to Right) *Acc+Gyro+EMG*, *Acc+Gyro* and *EMG* for data type (Top to Bottom) signal, image and audio respectively.

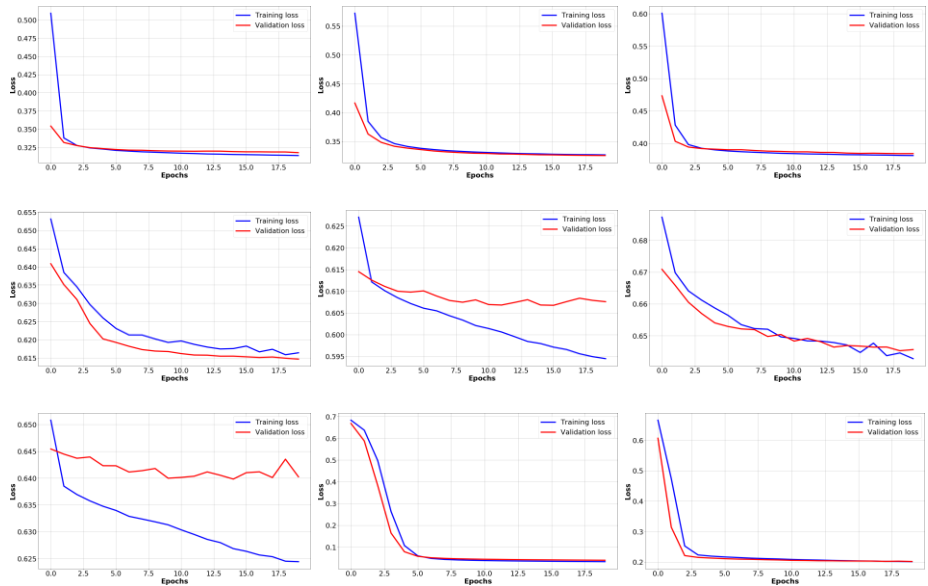

Fig. 12: DNN classification loss for datasets (Left to Right) Acc+Gyro+EMG, *Acc+Gyro* and *EMG* for data type (Top to Bottom) signal, image and audio.

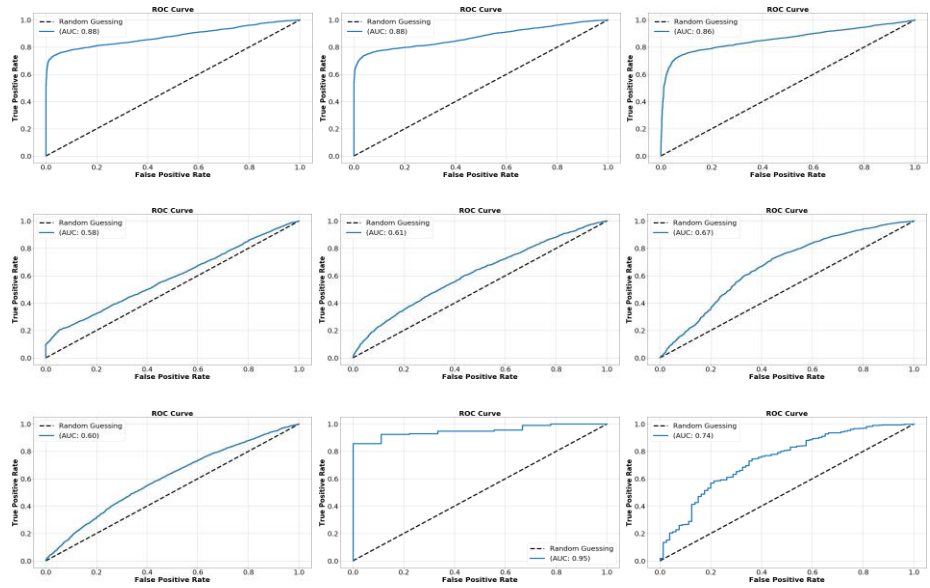

Fig. 13: DNN ROC curves for datasets (Left to Right) Acc+Gyro+EMG, Acc+Gyro and EMG for data type (Top to Bottom) signal, image and audio.

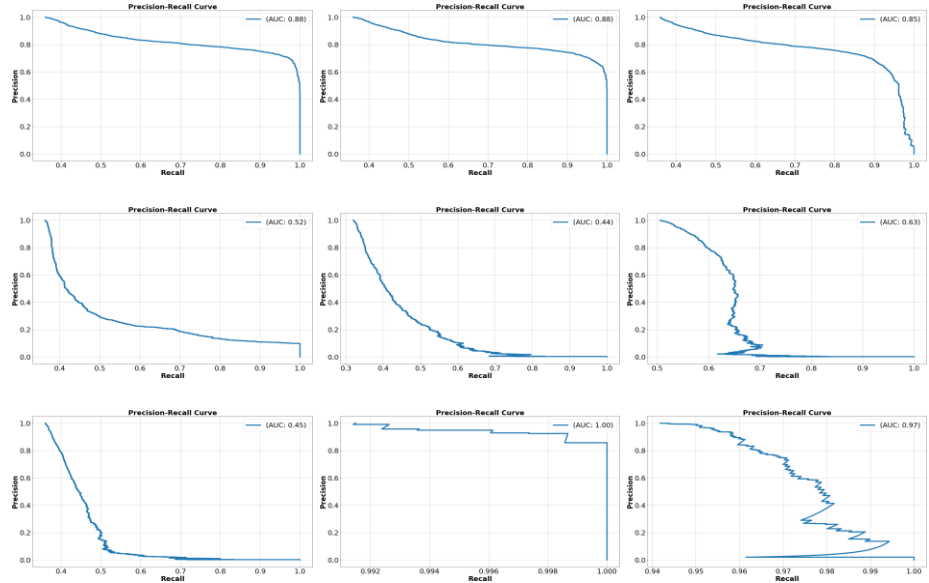

Fig. 14: DNN Precision-Recall curves for datasets (Left to Right) Acc+Gyro+EMG, Acc +Gyro and EMG for data type (Top to Bottom) signal, image and audio respectively.

## 7.2 Novelty Detection (Before and After Encoding)

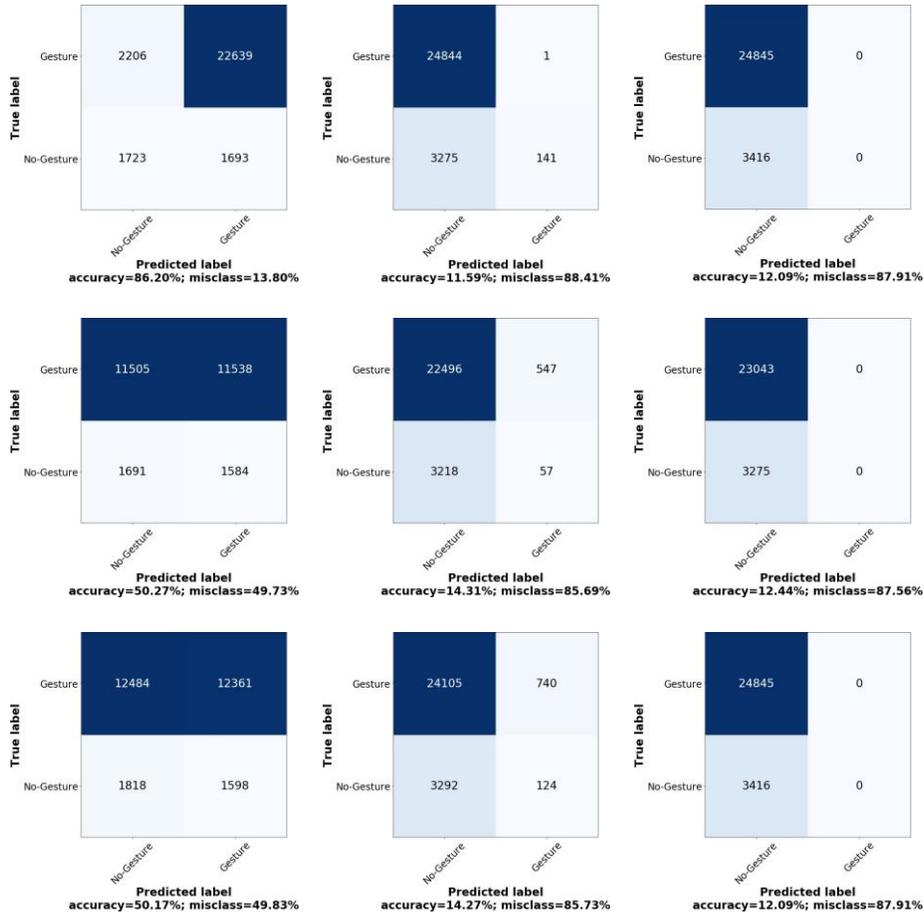

Fig. 15: **Consolidated data - (Acc+Gyro+EMG):** Novelty detection confusion matrices with models trained on 75% Gesture and tested on 25% *Gesture* + 100% *No-Gesture* instances for (Left to Right)*One-Class SVM, IsolationForest* and *GMM+Isotonic Regression* for data type (Top to Bottom) signal, image and audio respectively.

Table 4 summarize results from anomaly detection approaches described in Subsection 3.3. Models were trained on Gesture instances. Among all tested methods, on audio data for *Acc+Gyro, GMM+Isotonic Regression* achieves the highest accuracy of 96.12% with 96.12% precision and 100% sensitivity (bold gray area). *Isolation Forest* achieves the second highest accuracy, precision and sensitivity. However, One-Class SVM on signal data for *Acc + Gyro + EMG* and *Acc + Gyro* achieve highest specicity of 91.12% and 80.74% respectively.

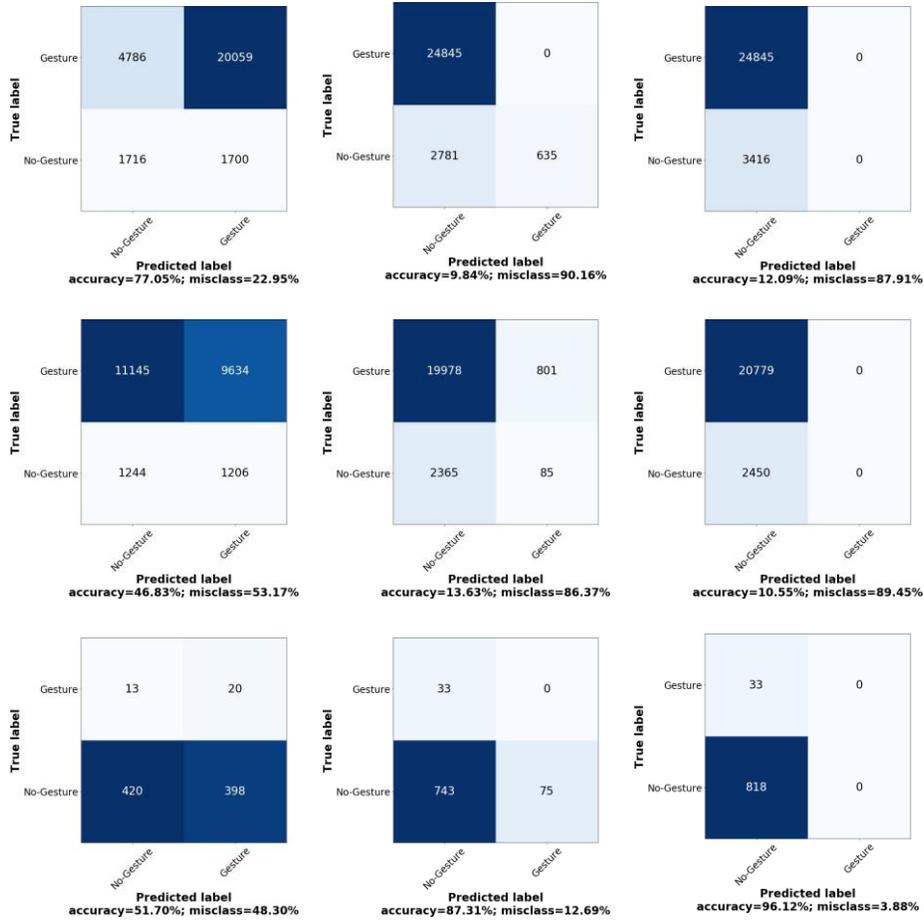

Fig. 16: <span style="color:blue">Consolidated data - (Acc+Gyro):</span> Novelty detection confusion matrices with models trained on 75% *Gesture* and tested on 25% *Gesture +* 100% *No-Gesture* instances for (Left to Right) *One-Class SVM, Isolation Forest* and *GMM+Isotonic Regression* for data type (Top to Bottom) signal, image and audio respectively.

From corresponding confusion matrices, for audio data, notice that *GMM + Isotonic Regression* detected all No-Gesture instance were detected correctly, but misclassified all *Gesture* instances. This is due to high class imbalance where we have fewer *Gesture* instances in comparison to *No-Gesture* ones. We notice an improvement in precision (97%) and specificity (60.61%) with *One-Class SVM* on audio data with *Acc + Gyro*, but lower accuracy (51.70%) and sensitivity (51.34%). Thus, with our proposed signal-audio encoding approach; we were able to detect novel class with higher precision while preserving most of the information in data transformation.

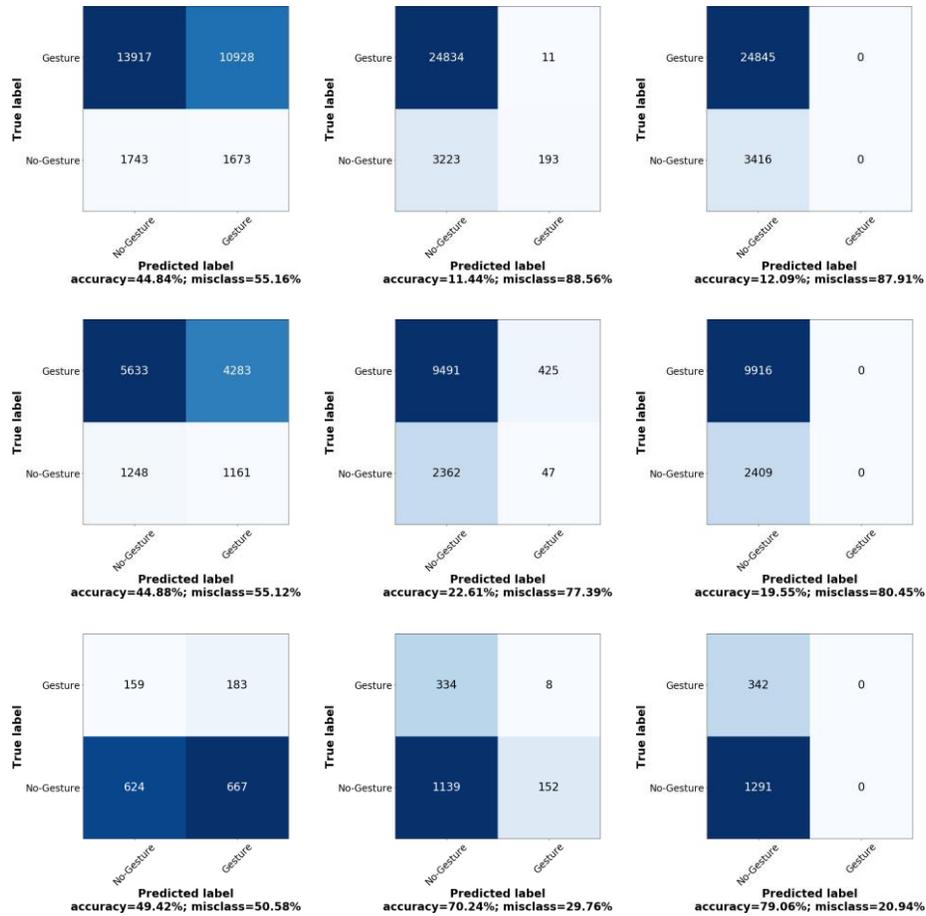

Fig. 17: **Dataset - EMG:** Novelty detection confusion matrices with models trained on 75% *Gesture* and tested on 25% *Gesture* + 100% *No- Gesture* instances for (Left to Right) *One-Class SVM, Isolation Forest* and *GMM+Isotonic Regression* for data type (Top to Bottom) signal, image and audio respectively.

Also, notice that all except *GMM + Isotonic Regression* anomaly detection methods took comparatively much less run time (of the order of a few seconds) than did other classical and deep learning methods (Table 5).

Table 4: Novelty detection results before and after encoding. Models were trained on *Gesture* instances as explained in Figures 15, 16, 17. Cells highlighted in gray color represent the models with highest precision. The best model is in bold.

| | Accuracy | Precision | Sensitivity | Specificity |
|---|---|---|---|---|
| **Before Encoding - Signal data (Acc + Gyro + EMG)** | | | | |
| One Class SVM | 86.20% | 43.85 | 50.44% | 91.12 % |
| Isolation Forest | 11.59% | 11.65 % | 95.87% | 0 |
| GMM with Isotonic Regression | 12.09% | 12.09 % | 100% | 0 |
| **Before Encoding - Signal data (Acc + Gyro)** | | | | |
| One Class SVM | 77.05% | 26.39 | 50.23% | 80.74 % |
| Isolation Forest | 9.84% | 10.07 % | 81.41% | 0 |
| GMM with Isotonic Regression | 12.09% | 12.09 % | 100% | 0 |
| **Before Encoding - Signal data (EMG)** | | | | |
| One Class SVM | 44.84% | 11.13 | 51.02% | 43.98 % |
| Isolation Forest | 11.44% | 11.49 % | 94.35% | 0.04 |
| GMM with Isotonic Regression | 12.09% | 12.09 % | 100% | 0 |
| **After Encoding - Image data (Acc + Gyro + EMG)** | | | | |
| One Class SVM | 50.27% | 12.81 | 51.63% | 50.07 % |
| Isolation Forest | 14.31% | 12.51 % | 98.26% | 2.37 |
| GMM with Isotonic Regression | 12.44% | 12.44 % | 100% | 0 |
| **After Encoding - Image data (Acc + Gyro)** | | | | |
| One Class SVM | 46.83% | 10.04 | 50.78% | 46.36 % |
| Isolation Forest | 13.63% | 10.58 % | 96.53% | 3.85 |
| GMM with Isotonic Regression | 10.55% | 10.55 % | 100% | 0 |
| **After Encoding - Image data (EMG)** | | | | |
| One Class SVM | 44.88% | 18.14 | 51.81% | 43.19 % |
| Isolation Forest | 22.61% | 19.93 % | 98.05% | 4.29 |
| GMM with Isotonic Regression | 19.55% | 19.55 % | 100% | 0 |
| **After Encoding - Audio data (Acc + Gyro + EMG)** | | | | |
| One Class SVM | 50.17% | 12.71 | 53.22% | 49.75 % |
| Isolation Forest | 14.27% | 12.02 % | 96.37% | 2.98 |
| GMM with Isotonic Regression | 12.09% | 12.09 % | 100% | 0 |
| **After Encoding - Audio data (Acc + Gyro)** | | | | |
| One Class SVM | 51.70% | 97.00 | 51.34% | 60.61 % |
| Isolation Forest | 87.31% | 95.75% | 90.83% | 0 |
| GMM with Isotonic Regression | 96.12% | 96.12% | 100% | 0 |
| **After Encoding - Audio data (EMG)** | | | | |
| One Class SVM | 49.42% | 79.69 | 48.33% | 53.51% |
| Isolation Forest | 70.24% | 77.33% | 88.23% | 2.34 |
| GMM with Isotonic Regression | 79.06% | 79.06% | 100% | 0 |

Table 5: Comparison of models' performance

| Model | Data | Sensors | Platform | Time taken to run the model (Hours-Minutes-Seconds) |
|---|---|---|---|---|
| Transfer Learning | Images | | GPU | 1 : 24 : 37 |
| DNN | Signal | Acc + Gyro + EMG | GPU | 0 : 04 : 06 |
| | | Acc + Gyro | | 0 : 04 : 00 |
| | | EMG | | 0 : 04 : 03 |
| DNN | Image Features | Acc + Gyro + EMG | GPU | 0 : 03 : 55 |
| | | Acc + Gyro | | 0 : 03 : 19 |
| | | EMG | | 0 : 02 : 25 |
| DNN | Audio Features | Acc + Gyro + EMG | GPU | 0 : 03 : 58 |
| | | Acc + Gyro | | 0 : 00 : 55 |
| | | EMG | | 0 : 01 : 07 |
| CJSD, M-CJSD, RBF Kernels in SVM | Signal | Acc + Gyro + EMG | CPU | 09 : 01 : 23 |
| | | Acc + Gyro | | 08 : 18 : 43 |
| | | EMG | | 16 : 14 : 41 |
| CJSD, M-CJSD, RBF Kernels in SVM | Image Features | Acc + Gyro + EMG | CPU | 2 days 00 : 41 : 50 |
| | | Acc + Gyro | | 03 : 34 : 59 |
| | | EMG | | 03 : 47 : 19 |
| CJSD, M-CJSD, RBF Kernels in SVM | Audio Features | Acc + Gyro + EMG | CPU | 12 : 07 : 27 |
| | | Acc + Gyro | | 00 : 06 : 49 |
| | | EMG | | 00 : 16 : 23 |
| Anomaly Detection (One-class SVM, Isolation Forest, GMM) | Signal, Image, Audio | Acc + Gyro + EMG / Acc + Gyro / EMG | CPU | All models except GMM take less than a few seconds. GMM takes several minutes ($\leq$ 36 minutes). |

## 8  Conclusion

In this paper we made contributions concerning the challenges to analyze signal data originating from multiple sources. In particular, we adopted a multi- sensor data fusion approach using IMU and EMG sensors to leverage abundant, raw, digital information for strategic advantage that enables intelligent military decision-making. Impetus behind our research arises from potential use of IoT devices for gesture recognition in military's Multi Domain operations (MDO). We investigated the challenge of enabling an intelligent identification and detection operation and demonstrated the feasibility of the proposed Machine Learning, Deep Learning and Anomaly Detection models to support a detection and identification of hand gestures.

Moreover, in order to leverage communication in C2 (Command and Control) systems, and limitations of Blockchain technology which is commonly used to facilitate secure sharing of IoT datasets, we introduced data encoding approaches to modify information (signal) by transforming it to an image and audio signal which are invertible and easier to visualize. We proposed zero padding to ad- dress the challenges of very short signals in encoding. In general, our encoding approach is applicable to military environments where IoT devices maybe used for secure data transmission.

We evaluated all methods on hand gesture datasets before and after encoding. Results from anomaly detection with GMM + Isotonic Regression confirmed that it achieved higher accuracy on audio data while One-Class SVM achieved higher precision over other methods on audio data. Also, classification methods achieved higher accuracies on audio data. Thus, our proposed encoding approach preserved most of the information in data transformation. Another virtue of tested anomaly detection methods is extremely low computational complexity in terms of time and memory over other classification approaches.

Future work will extend our analysis to a more complex heterogeneous IoT environment. We will investigate the impact of governing parameters (IoT topology, connectivity, security, etc.) on model performance for data fusion. Although, our analysis on data before and after encoding showed promising results, future work will investigate data regularities and how the data patterns are impacted.

**Acknowledgments** Author will like to thank to Dr. Mark Dennison for providing *Gesture Myo Dataset* and explaining data collection process.